\documentclass[11pt]{article}

\usepackage[final]{acl}
\usepackage[ragged]{footmisc}

\usepackage{times}
\usepackage{latexsym}
\usepackage{tipa}

\usepackage[T1]{fontenc}

\usepackage[utf8]{inputenc}

\usepackage{microtype}

\usepackage{inconsolata}
\usepackage{upquote}

\usepackage{graphicx}

\title{Some Dialects Are More Equal Than Others: \\Non-Prestigious Arabic Dialectal Bias in LLMs}

\author{
 \textbf{Mai Mohamed Eida\textsuperscript{1}},
 \textbf{Ryan Dolan\textsuperscript{1}},
 \textbf{Paul de Nijs\textsuperscript{1}},
 \textbf{Jonathan Dunn\textsuperscript{1}}
\\
 \textsuperscript{1}University of Illinois Urbana Champaign
 \\
  \texttt{\{maimm2, ryanpd3, pdenijs2, jedunn\}@illinois.edu}
}

\begin{document}
\maketitle
\makeatletter
\renewenvironment{abstract}%
  {\begin{center}\large\textbf{\abstractname}\end{center}%
    \begin{list}{}%
      {\setlength{\rightmargin}{0.6cm}%
        \setlength{\leftmargin}{0.6cm}}%
      \item[]\ignorespaces%
      \@setsize\normalsize{12pt}\xpt\@xpt
  }%
  {\unskip\end{list}}
\makeatother
\begin{abstract}
Previous work on Egyptian Arabic in NLP has focused largely on the prestigious Cairene Egyptian Arabic (CEA) dialect, resulting in a lack of representation for the less prestigious Sa'idi Egyptian Arabic (SEA) dialect both in LLM and resource development. Does this lack of representation influence an LLM's view of the acceptability of SEA (upstream), and does an upstream bias against SEA lead to worse performance (downstream)? We investigate the upstream effect of SEA dialectal features on LLM preferences in a Targeted Syntactic Evaluation (TSE) task which reveals a significant bias against SEA across multiple LLMs. We then analyze the effect of these same features on downstream model performance on MMLU benchmarks and show that models experience a degradation in performance when presented with SEA. This work highlights the need for further exploration on how sub-dialectal variation impacts language technologies.
\end{abstract}
\section{Introduction}
Arabic dialectal variation remains a continuous challenge in Arabic Natural Language Processing (NLP). With over twenty-six dialects, and hundreds of sub-dialects~(\citealp{Behnstedt:2013:AD};~\citealp{versteegh:2006:EALL}), a central goal of the field has been to model dialectal varieties of Arabic across language technologies, including Large Language Models (LLMs). However, accounting for such diverse dialectal variation is difficult and does not necessarily account for sub-dialectal variation within each country, and therefore, in this process, much of sub-dialectal variation is lost in practice. 

Egyptian Arabic (EA) provides a clear case for supporting marginalized sub-dialectal varieties in Arabic language technologies. Egyptian Arabic is the most thoroughly modeled dialect in Arabic NLP, but it has tacitly become synonymous with ``Educated Cairene Egyptian Arabic'', the most prestigious dialect \citep{bassiouney:2018:CTS} spoken by middle class educated Egyptians in Cairo and surrounding urban cities~(\citealp{abdel-malek:1972:TCC};~\citealp{Badawi:1973:MAA}). This variety represents only one of the five groups of Egyptian Arabic sub-dialects, which vary across phonology, morphology, syntax, semantics, and lexicon~(\citealp{behnstedt:1985:DAA};~\citealp{Badawi:1973:MAA}) and also vary culturally from Cairo. In this paper, we focus on the second most spoken dialect in Egypt, Sa'idi Egyptian Arabic (SEA) which is not linguistically represented within the dialectal landscape in current Arabic language technologies~\cite{Eida:2024:HWT}. 

The focus on Cairene Egyptian Arabic (CEA) is consistent with accessibility and availability of resources for prestigious dialects. Dialect attitudes are mediated by socioeconomic status, educational access, and language ideology~\cite{labov:2006:TSS}. Speakers of prestigious dialects have the means to participate in digital spaces and contribute linguistic data which increases their representation in language technologies. On the other hand, speakers of marginalized dialects may avoid using their dialects online, even when access is available. This produces a self-reinforcing cycle in which language technologies prefer certain dialects while excluding others due to lack of linguistic resources. 

Previous work has demonstrated that SEA is not well represented in online Egyptian Arabic corpora~\citep{Eida:2024:HWT}. However, whether this lack of representation actually results in upstream or downstream bias against SEA in model behavior has not yet been demonstrated. It is possible that, despite the scarcity of SEA representation in online corpora on which LLMs are trained, the internal representations learned by LLMs are generalized enough that they are insensitive to dialect differences. If so, models may exhibit comparable downstream performance on SEA and CEA, suggesting that the scarcity of SEA resources does not necessarily translate into a poorer experience for SEA speakers.

This paper addresses two research questions. \textbf{First, do LLMs replicate the same bias humans have towards prestigious and non-prestigious sub-dialects in their upstream representations?} If so, this supports previous work on the lack of SEA representation in pre-training data in Arabic NLP and that such lack of representation can result in an upstream bias. In experiment 1, our results confirm there is a bias in monolingual, bilingual, and multilingual LLMs against SEA, with internal representations favoring CEA or Modern Standard Arabic (MSA) over SEA in almost every single case. \textbf{Our second research question addresses the implications of upstream bias: does this upstream bias impact the performance of LLMs on downstream tasks?} Specifically, we examine performance on Arabic Massive Multitask Language Understanding (MMLU) benchmarks. If there is no impact on downstream tasks, this may shift the focus of future work to other necessary tasks such as cultural awareness of sub-dialectal variation rather than curating SEA-specific corpora. The results from our second experiment show that downstream bias varies more across models, but still shows consistent favor towards CEA overall\footnote{All data, code, and prompts are available on \href{https://github.com/maimm2/EA_Dialectal_Bias}{https://github.com/maimm2/EA\_Dialectal\_Bias}.}.

\section{Background}
\subsection{Language Attitudes}
Language attitudes are structured by social prestige and power~\cite{labov:2006:TSS}. Dialect choice is not neutral. It is shaped by socially embedded evaluations that influence both perception and production. Speakers converge toward or diverge from others’ speech depending on social attitudes and group identity~\cite{giles:1970:ERT}. Perceptual dialectology shows that listeners systematically associate dialects with social attributes, assigning traits such as intelligence, competence, and professionalism to prestigious varieties, while attributing negative characteristics such as lack of education or credibility to marginalized dialects~(\citealp[]{giles:1970:ERT};~\citealp{lambert:1960:ERT};~\citealp{Preston:1993:FD}).

This extends to Arabic dialects, with language attitudes spanning sub-dialects of many Arabic dialects (\citealp{aldosaree:2016:LAT}; \citealp{eltouhamy:2015:LAT}; \citealp{hachimi:2015:GABA}). As the most prestigious dialect, Cairene Egyptian Arabic (CEA) is socially constructed as educated, modern, and authoritative, while Sa'idi Egyptian Arabic is stereotyped as rural, uneducated, and socially marginal~(\citealp{bassiouney:2017:IAD}; \citealp{haeri:1997:TSM}). This bias affects how speakers are evaluated and treated in education, employment, and institutional settings, with speakers of prestigious dialects being perceived as more competent and credible than speakers of stigmatized varieties. However, do LLMs exhibit a similar bias toward prestigious dialects mirroring existing human bias?

\subsection{Dialectal Bias in LLMs}
Recent research has shown that LLMs frequently demonstrate bias against certain language varieties across many different languages. \citet{lin:2025:ADF} show that LLMs systematically display a high degree of unfairness (performance gaps) and brittleness (sensitivity to input perturbations) when responding to reasoning queries in African American English (AAE). Similar work by \citet{Pan:2025:ADB} shows that such bias also exists for multiple-choice reasoning benchmarks and extends to other English varieties such as Chicano, Appalachian, Southern, Indian, and Singaporean English. These works crucially demonstrate that even in tasks such as code generation or queries for factual world knowledge which should not reasonably vary across dialects, LLMs show a systematic bias.

Outside of benchmark performance bias, \citet{fleisig:2024:LBI} use qualitative evaluations from native speakers of 10 different varieties of English to show that GPT-3.5 and GPT-4 often use demeaning and stereotypical language when responding to varieties other than Standard American and Standard British English. They show that the models often misuse or exaggerate dialectal features in a way that is unnatural and perceived as mockery. \citet{Hassan:2025:DPB} additionally show that LLMs demonstrate a systematic bias against AAE when used for sentiment analysis tasks, with AAE language being much more likely to be identified as negative. Outside of English, \citet{Bui:2025:LLM} show that systematic bias also exists for non-standard dialects of German such as Bavarian and Alemannic varieties, with non-standard varieties more likely to be associated with negative adjectives in an adjective-association task. There is also work on Bengali and dialectal bias in this low-resource language based on the minorities/power structures~\cite{sadhu:2025:SBI}, showing LLM production exhibit significant gender and religious bias in Bangala. In all of this research, LLMs do exhibit bias against the less prestigious varieties, and this impacts downstream performance. Is this also the case for Egyptian Arabic?

\section{Experiment 1: Do LLMs have a preference for one dialect?}
Since SEA is highly under-represented compared to CEA in existing Egyptian Arabic datasets and tools~\cite{Eida:2024:HWT}, 
is there a preference for one dialect over the other due to the scarcity of data? If so, then we expect LLMs to prefer CEA to SEA by finding CEA sentences more probable than an equivalent SEA sentence in a minimal pair.

\subsection{Methodology}
\paragraph{Models} We select five pretrained models to test upstream preferences: one monolingual Arabic model specifically trained on both MSA and Tweets in Dialectal Arabic (DA) (henceforth ``monolingual'', Jasmine-350M: ~\citealt{billah:2023:jasmine}), two English-Arabic bilingual models trained on a variety of Arabic dialects (henceforth ``bilingual'', Jais-2-8B \& Jais-1-13B: \citealt{Anwar:2025:JAIS2};~\citealt{sengupta:2023:JAIS13b}), one multilingual model trained on generic Arabic data (henceforth ``multilingual'', Llama 3.1-8B:~\citealt{llama:3model:card}), and one multilingual model continuously-pretrained specifically on Egyptian Arabic (henceforth ``EA'', NileChat-4B:~\citealt{shang:2025:NILECHAT})\footnote{NileChat-4B is only continuously-pretrained on Egyptian Arabic text, but is later instruct-tuned with both Egyptian Arabic and English.}.

\paragraph{Task} To test upstream preferences, we employ Targeted Syntactic Evaluation (TSE, \citealp{linzen:2016:ATA}). In this approach, the probabilities of two sentences in a minimal pair are compared to decide which of the minimal pair of dialects the model prefers. We create minimal pairs between CEA and SEA by sampling items from an SEA corpus \citep{eida:2025:BCS} and changing features specific to SEA to their CEA equivalents. These features include phonetic, morphological, orthographic, and lexical variations. For most minimal pairs, only one feature at a time is changed, although we also include a ``mixed'' condition in which multiple features are changed at once.
\begin{figure}
\centering
\includegraphics[width=1\linewidth]{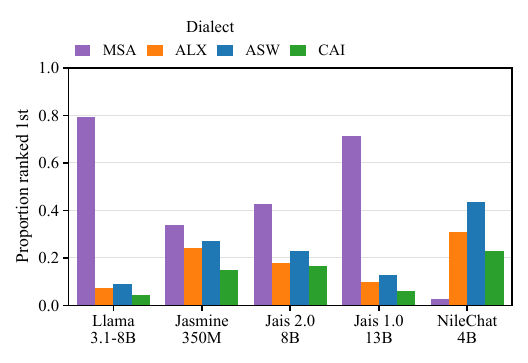}
\caption{Proportion of dialect preference by model in MADAR, defined as being the most preferred variation in the TSE task. MSA is consistently most preferred, followed ASW (Aswan), ALX (Alexandria), then CAI (Cairo). With the exception of NileChat-4B, MSA is most preferred. NileChat-4B was precontinuously trained on Egyptian Arabic and therefore, preferring EA to MSA is expected.}
\label{fig:preferredminimalpairscountsMADAR}
\end{figure}
\paragraph{Minimal Pair Construction} We additionally create an ungrammatical baseline (UG) in where we replace the valid Arabic features such as morphemes and lexical items with meaningless nonce forms that are less likely to correspond to an Arabic variety. For example, the CEA future marker h and its SEA counterpart \textrevglotstop{} were replaced with d, which does not function as a future tense marker in Arabic dialects. We consistently replaced d as future tense in every applicable case thereafter. This ensures that the modified forms preserved superficial characteristics such as length while removing their grammatical interpretation, and we control for as much variable change as possible. The UG forms generated matched the morphological, phonological, and orthographic features in length and script, as illustrated in Appendix~\ref{samples} in Table~\ref{tab:listoffeatueresacrossbenchmarks}. Lexical features were altered with +/-4 letters depending on how easy it was to create a nonce word. This accounts for the possibility of an LLM considering one or two letter changes in variants a typological error (``typo''). The UG baseline allows us to identify whether the model prefers CEA over SEA because it finds SEA to be purely ungrammatical, or because it considers SEA as a valid but dispreferred dialect. This yields a total of 134 minimal pairs, with samples illustrated in Table~\ref{tab:minimalpairssamplesandstats} in Appendix~\ref{samples}.

\paragraph{Evaluation} To compute sentence probabilities, we follow \citet{leong:2023:LMC} in our implementation of TSE by adding the (negative) log probabilities of each token in a given sentence (conditioned on previous tokens) to obtain the model's ``judgment'' for that sentence. We then normalize this value by the length of the sequence to ensure that longer sentences are not artificially less probable.

As MSA remains the dominant Arabic variety in available NLP resources and is substantially better represented than dialectal Arabic in LLM training data~\cite{mousi:2025:ARADICE}, we first check if MSA is the most preferred dialect over Egyptian Arabic (EA). To do so, we use the MADAR~\cite{bouamor:2018:MADAR} parallel Arabic corpus and extract Arabic minimal pairs between MSA and the three Egyptian cities in the corpus: Cairo (CAI), Alexandria (ALX), and Aswan (ASW). We hypothesize that MSA will be the most preferred variety, given both its prestige and prevalence in available training data. Then, we examine which of our three CEA, SEA, and UG variants is most preferred by the model in each minimal pair. We then determine pairwise preferences between CEA and SEA and SEA and the ungrammatical baseline (UG). If the model exhibits a bias against SEA, we expect that the CEA variant will be consistently ranked the most probable sentence. If little or no bias exists, we expect that CEA will be ranked over SEA in roughly equal proportion to SEA being ranked over CEA. Furthermore, if CEA is consistently ranked higher than SEA, and SEA is consistently ranked higher than UG, this would indicate that the model recognizes SEA as grammatical, but still disprefers it to CEA. If there is no consistent preference for SEA over UG, this indicates that the model views SEA as fully ungrammatical.
\begin{figure}
\centering
\includegraphics[width=1\linewidth]{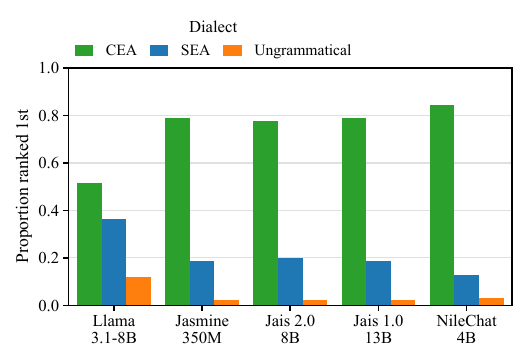}
\caption{Proportion of dialect preference by model, defined as being the most preferred variation in the TSE task. CEA is consistently most preferred, followed by SEA and UG.}
\label{fig:preferredminimalpairscounts}
\end{figure}
\begin{figure}[]
\centering
\includegraphics[width=1\linewidth]{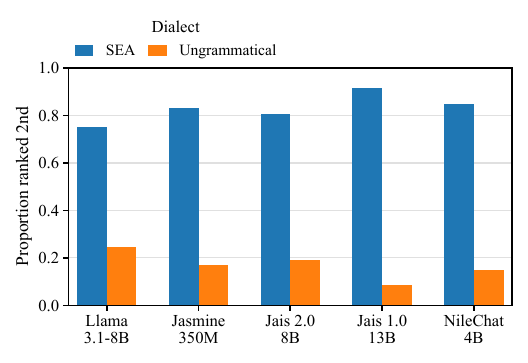}
\caption{Second-choice model preferences when CEA is ranked most probable. All models strongly prefer SEA to the Ungrammatical baseline, indicating that SEA is viewed as legitimate, but dispreferred, variation.}
\label{fig:preferredminimalpairscounts_secondplace}
\end{figure}
\subsection{Results}

\textbf{Is MSA more preferred than EA?} As illustrated in Figure~\ref{fig:preferredminimalpairscountsMADAR}, MSA is the most preferred variety of Arabic when compared to Egyptian Arabic. Therefore, our hypothesis that MSA, as both the most prestigious and the most represented variety in training data, would be preferred holds across all models except NileChat-4B, where Egyptian Arabic emerges as the most preferred variety. Given that NileChat-4B is the only model continuously pre-trained specifically on Egyptian Arabic, its preference for Egyptian Arabic over MSA is expected.

\textbf{Do LLMs prefer CEA over SEA?}
The ranking preferences of each model are given in Figure~\ref{fig:preferredminimalpairscounts}. In all models, a clear preference is shown for the prestigious CEA dialect over the less prestigious SEA dialect. The preference is very clear in the Arabic monolingual (Jasmine-350M), Arabic-English bilingual (Jais-1-13B and Jais-2-8B), and Egyptian Arabic (EA) model (NileChat-4B), with CEA being preferred around 80\% of the time. This preference notably drops in the multilingual model (Llama-3.1-8B) to only 50\%. A similar pattern is found for the ungrammatical baseline UG, which is found most preferable only around 2\% of the time for all models except Llama-3.1-8B (11\%). Therefore, LLMs prefer CEA over SEA and display an upstream bias against SEA. 

\textbf{Do LLMs consider SEA a legitimate variety?} Having demonstrated that all models prefer CEA to SEA, we now ask whether this is because SEA is viewed as completely ungrammatical, or whether the model recognizes the SEA variants as being legitimate, but still dispreferred. The results in Figure~\ref{fig:preferredminimalpairscounts} suggest that SEA should be viewed as legitimate given that it is ranked above CEA in a non-negligible number of cases. Indeed, we find that in cases where CEA is ranked to be the most probable, SEA is preferred over the ungrammatical baseline in around 90\% of cases (Figure ~\ref{fig:preferredminimalpairscounts_secondplace}). Thus, it is not the case that SEA is viewed as completely ungrammatical; rather, the models are simply biased towards CEA.

The absolute rankings of negative log probabilities clearly demonstrate a preference hierarchy of CEA > SEA > UG across all models, with only Llama-3.1-8B showing a somewhat weaker preference. Thus, in the majority of cases, CEA is indeed preferred over SEA, but SEA dialectal features are recognizably legitimate variations such that SEA is still systematically preferred over the UG ungrammatical baseline.

\textbf{To what degree do LLMs prefer SEA over the ungrammatical baseline?} Having established this preference hierarchy and the fact that SEA is not considered purely ungrammatical, we now turn to the question of whether SEA is considered closer to CEA overall than to the ungrammatical baseline. We do so by comparing the distribution of the \textit{differences} in sequence probabilities between CEA and SEA ($P(Seq_{CEA}) - P(Seq_{SEA})$) and SEA and the ungrammatical baseline UG ($P(Seq_{SEA}) - P(Seq_{UG})$) for each minimal pair using a paired $t$-test. This allows us to determine the degree of preference; if, on average, $P(Seq_{CEA}) - P(Seq_{SEA}) < P(Seq_{SEA}) - P(Seq_{UG})$, this implies that the model considers SEA to be overall closer to CEA than the ungrammatical baseline. If the converse is true, this implies that the model considers SEA to be closer to the ungrammatical baseline than to CEA, even if it consistently ranks SEA higher.

The paired $t$-test is only significant for Jais-1-13B ($t = -2.59, p = 0.03^*$), which indicates that for all other models, SEA is not closer on average to CEA than to complete ungrammaticality, or vice versa. This can also be seen visually in appendix ~\ref{ProbabilityDistributions}. This shows that SEA interestingly occupies a sort of midpoint between CEA and UG in terms of grammaticality.

\textbf{Does the number of characters changed matter?} Finally, we assess whether the upstream bias exhibited by the models correlates with the scale of the grammatical feature change as measured by character edit distance. We correlate minimum edit distance and the difference in probabilities examined in the previous section. Edit distance was positively correlated and statistically significant with the probability difference CEA - SEA $(r=0.18, p\approx 0.00)$ and CEA - UG $(r=0.17, p \approx 0.00)$ for all models except for Llama-3.1-8B. However, SEA - UG did not show a significant correlation $(r=0.009, p=0.8)$ with edit distance for any model. This shows that the magnitude of the change does affect the degree to which CEA is preferred over SEA, but that UG is heavily dispreferred regardless of the number of characters changed.

In summary, the consistent ranking of CEA over SEA shows a strong bias against SEA in upstream model preferences, answering our first research question. It is especially noteworthy that the minimal pairs were constructed using samples from an SEA corpus, changing only one overt SEA feature to its CEA counterpart. Despite this SEA source context, models still consistently prefer the CEA variant once that single feature is changed. In other words, the models prefer a CEA feature in an SEA context even over an SEA feature in an otherwise SEA context. Thus, the preference is strong enough that the model prefers an out-of-place prestigious dialectal feature to a well-suited non-prestigious dialectal feature. On the other hand, SEA is ranked higher than the ungrammatical baseline, demonstrating that SEA is a viable but dispreferred dialect occupying a midpoint between the prestigious CEA dialect and total ungrammaticality.

\section{Experiment 2: Does Upstream Bias Impact Downstream Performance?}
The results in Experiment 1 reveal a persistent upstream dialectal preference for CEA over SEA. However, does upstream bias actually impact downstream performance? Experiment 2 aims to answer this question. If so, this would indicate that marked sub-dialectal variation affects downstream performance, and emphasizes the need for sub-dialectal variation in resource development, model development, and language technologies overall.

\subsection{Methodology}
We adapt~\citet{Pan:2025:ADB}'s methodology, which evaluates the extent to which grammatical features associated with non-standard English varieties influence LLM performance on benchmarks such as Massive Multitask Language Understanding (MMLU), Boolean Questions (BoolQ), and Science Questions (SciQ). We extend this methodology to three Arabic varieties: MSA, CEA, and SEA using the same set of features as in Experiment 1. We exclude lexical and orthographic variations because they are more difficult to consistently apply across all benchmarks. For each feature, we use a regular expression to first select all benchmark questions in which the feature occurs. We then create a variant of the question and multiple-choice answers with the target feature changed to its equivalent in MSA, CEA, and SEA. As in Experiment 1, we also produce an ungrammatical control variant (UG) as a baseline following the same rules of length, script and consistency. This transforms each original item into four versions corresponding to MSA (ArabicMMLU only), CEA, SEA, and UG conditions.

\paragraph{Benchmarks} For this experiment, we focus on three MMLU benchmarks specifically developed for Arabic varieties. We select one MSA MMLU benchmark, ArabicMMLU~\cite{Koto-etal:2024:AMA}, and two Dialectal Arabic (DA) benchmarks, AraDiCE-ArabicMMLU-EGY~\cite{mousi:2025:ARADICE}  and DialectalArabicMMLU EGY~\cite{altakrori:2026:DAMMLU}. For the latter two benchmarks, we extract the Egyptian Arabic items only. 

\begin{table*}[ht]
\centering
\includegraphics[width=1\linewidth]{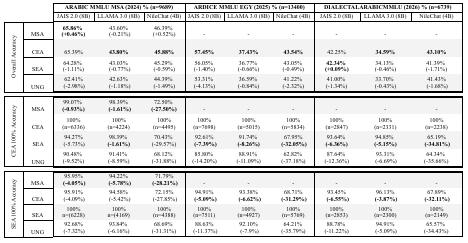}
\caption{\textbf{Top:} Overall accuracy on single grammatical feature minimal pairs. \textbf{Middle:} Accuracy on the corresponding MSA, SEA, and ungrammatical (UG) variants restricted to items answered correctly in the CEA condition (100\% CEA subset). \textbf{Bottom:} Accuracy on the corresponding MSA, SEA, and ungrammatical (UG)  variants restricted to items answered correctly in the SEA condition (100\% SEA subset).  Unperturbed items are excluded from the overall evaluation. Random-chance accuracy is 29\% for ArabicMMLU and AraDiCE-ArabicMMLU-EGY, and 25\% for DialectalArabicMMLU. Bold indicates the highest accuracy within each model and benchmark column. Values in parentheses indicate the absolute percentage-point difference relative to the corresponding CEA-correct or SEA-correct baseline performance.}
\label{tab:accuracypertubatedscoresisolated}
\end{table*}
This creates two conditions: one in which the original questions are fully in MSA, and one in which the original questions are in Egyptian Arabic. Having both types of conditions is important, because the model may behave very differently when faced with an SEA feature in MSA versus an SEA feature in Egyptian Arabic.  Additionally, because MSA is a standard/neutral variety and the most preferred compared to EA as illustrated in Experiment 1, we can safely add CEA and SEA features in isolation from one another without the potential confound of an item already being written in CEA. By using MSA as the source condition in at least one benchmark, we can more directly isolate the effect of introducing CEA and SEA features while holding all other aspects of the item constant. The two Egyptian Arabic benchmarks are still included because both CEA and SEA are primarily spoken dialects, and thus provide a more natural context in which SEA features would actually occur. We list the benchmarks and the 24 extracted features in Table~\ref{tab:listoffeatueresacrossbenchmarks} in Appendix~\ref{samples}.

\paragraph{Models} We evaluate benchmark performance on three of the five models examined in Experiment 1: Llama 3.1-8B \cite{llama:3model:card}, Jais-2-8B \cite{Anwar:2025:JAIS2}, and NileChat-4B \cite{shang:2025:NILECHAT}. Jasmine-350M \cite{billah:2023:jasmine} was excluded because it was not instruct-tuned, and Jais-1-13B \cite{sengupta:2023:JAIS13b} was excluded because the results of Experiment 1 were extremely similar to those of the newer Jais-2-8B.

\paragraph{Evaluation} We initially evaluate each grammatical feature in isolation, modifying a single feature at a time and measuring benchmark accuracy across the MSA, CEA, SEA, and UG conditions. We then introduce a ``mixed-features'' condition, in which two or more grammatical features are changed simultaneously within the same benchmark item. Some feature perturbations involve only minimal changes, sometimes only a single character. It is possible that models treat these forms as spelling variations or noise rather than meaningful dialectal distinctions. This is corroborated by the positive correlation between edit distance and difference in upstream preference in Experiment 1 between CEA and SEA. Therefore, evaluating both isolated and combined feature substitutions we can determine whether minimal dialectal perturbations affect model accuracy to the same degree as larger dialectal perturbations. 
\begin{table*}[]
\centering
\includegraphics[width=1\linewidth]{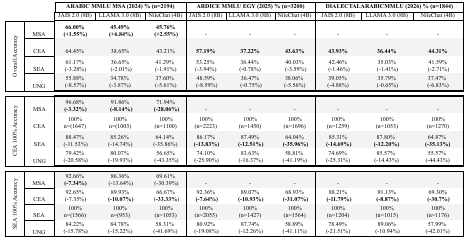}
\caption{\textbf{Top:} Overall accuracy on 2+ mixed grammatical feature minimal pairs. \textbf{Middle:} Accuracy on the corresponding MSA, SEA, and ungrammatical (UG) variants restricted to items answered correctly in the CEA condition (100\% CEA subset). \textbf{Bottom:} Accuracy on the corresponding MSA, SEA, and ungrammatical (UG) variants restricted to items answered correctly in the SEA condition (100\% SEA subset). Unperturbed items are excluded from the overall evaluation. Random-chance accuracy is 29\% for ArabicMMLU and AraDiCE-ArabicMMLU-EGY, and 25\% for DialectalArabicMMLU. Bold indicates the highest accuracy within each model and benchmark column. Values in parentheses indicate the absolute percentage-point difference relative to the corresponding CEA-correct and SEA-correct performance.}
\label{tab:accuracypertubatedscoresmixed}
\end{table*}

\section{Results}
\paragraph{Single Feature Accuracy Overall}
The results for the single feature condition are given in Table ~\ref{tab:accuracypertubatedscoresisolated}. We first report the accuracy across all items on each benchmark for each model and dialect condition (``Overall Accuracy''). The results show that across almost all models and benchmarks, the model achieves higher accuracy when the questions and answers are in CEA than when they are in SEA. The magnitude of difference is not particularly consistent across models, with different models being impacted more across different benchmarks. Furthermore, the original dialect of the question does not appear to impact the degree of difference, with ArabicMMLU (MSA) having a similar margin of difference as compared to AraDiCE-ArabicMMLU-EGY and DialectalArabicMMLU (Egyptian Arabic).

Although higher accuracy is observed on CEA as compared to SEA, the margin is often small (and in fact there is one marginally positive case in favor of SEA: Jais-2-8B in the DialectalArabicMMLU benchmark). This is consistent with the results of~\citet{Pan:2025:ADB}, who observe that non-prestigious English varieties degrade performance at a similarly small margin (between -0.5\% and -5\%). However, it is important to note that finding downstream performance degradation overall (though marginal) in favor of CEA is still noteworthy given that we focus on a single grammatical perturbation at a time, and do not include any syntactic, lexical or orthographic changes in the perturbations. This could indicate that more SEA linguistic dialect markers could contribute in further performance degradation in downstream performance when the full dialect is used. Another possible reason for the marginal difference is that benchmark performance is generally quite low across most models, peaking at around 66\% in the best case. Thus, it may be that many cases in which the model answers a question in SEA incorrectly can be attributed to general benchmark difficulty, and not difficulty specific to understanding the dialect. We therefore follow \citet{Pan:2025:ADB} in also examining the difference in accuracy on the questions which the model got correct in the preferred dialect (in this case, CEA). In other words, if the model got the question correct in the preferred dialect, how likely is the model to get that same question wrong if SEA is used instead and vice versa?

\paragraph{Single Feature Accuracy - CEA/SEA Correct Only}
Examining only the questions which the model got correct in CEA, the difference between CEA and SEA becomes much more severe. NileChat-4B in particular exhibits an extremely strong effect, with performance dropping nearly 35\% when the question is posed in SEA. Llama-3.1-8B by comparison shows the smallest differences almost across the board, which is consistent with its comparative lack of upstream bias observed in Experiment 1. However, these differences are certainly not negligible, with performance degrading up to 8.3\% on questions that were answered correctly in CEA. Lastly, Jais-2-8B exists at somewhat of a midpoint, sometimes patterning closely to Llama-3.1-8B and other times showing a stronger bias, but still far less than NileChat-4B. This shows that having an extremely strong upstream preference does not necessarily lead to a proportional impact on downstream performance.

To ensure the robustness of our results, we extend \citet{Pan:2025:ADB}'s approach looking at prestigious-correct only items to also examining items which the model got correct in the non-preferred dialect (SEA) as well. Examining both CEA-correct and SEA-correct items ensures that our analysis does not presuppose that there are no items that the model got correct in CEA but not SEA. We observe in Table~\ref{tab:accuracypertubatedscoresisolated} that the absolute magnitudes of the number of questions got correct on each benchmark is quite similar between both CEA and SEA. It will therefore be useful to validate the systematicity of the downstream bias with statistical means, since it not clear from the counts/proportions alone that the bias is significant. We will return to this point after reviewing the mixed-feature results.

\paragraph{Mixed-Feature Accuracy} Turning to the mixed feature conditions (Table ~\ref{tab:accuracypertubatedscoresmixed}), we observe that many of the same patterns hold, but are more pronounced. Both the overall and CEA-correct accuracies are lower in the mixed feature condition than in the single feature condition across all models and benchmarks. Therefore, as more features associated with SEA are introduced into the prompt, the more likely the model is to get the question wrong. This could be for a few different reasons: one reason is simply that the more dialectal features are introduced into the prompt, the more probable it is that the model will encounter a feature with which it is less familiar. Another possibility is that introducing more features makes it more likely that a feature which materially affects the question/answers is changed. For example, changing the plurality of a single noun in the question may not affect the model's ability to recognize the overall meaning of the prompt, whereas replacing a negation marker with one which is no longer comprehensible to the model may impact model performance significantly.

\paragraph{Which models have a significant bias against SEA in downstream performance?} We have so far demonstrated that across nearly all models and benchmarks, questions posed to the model in CEA are more likely to be answered correctly than those in SEA. However, the margin of difference for many models and benchmarks is marginal, sometimes within one percent. Moreover, the CEA-correct patterns mirror that of SEA-correct patterns, and therefore, how do we know this is attributed to dialect? To determine which models have significant bias against SEA in downstream performance, we perform McNemar's test between each pair of dialects, treating each question as a sample and whether the question was answered correctly by the model as the response variable \citep{Pan:2025:ADB}. We perform the paired tests independently for each model, benchmark, and number of features changed (single vs. mixed). If the result of the test is significant for that pair of dialects, this substantiates that the apparent presence of bias found in the accuracy score differential is not due to chance.

The results show that across nearly all benchmarks, Jais-2-8B exhibits the most consistent bias, with the proportion of correct answers in CEA being significantly higher than those in SEA. The only exception is in the DialectalArabicMMLU benchmark, in which it only becomes significant after three or more SEA features are present. NileChat-4B exhibits a significant bias in favor of CEA on DialectalArabicMMLU, as well as a significant bias in the case where two or more SEA features are present, yet the effect is no longer significant when three or more features are present. In all other benchmarks, the presence of SEA features is not significantly correlated with lower performance on the benchmark.  Llama-3.1-8B exhibits the least significant downstream bias, with SEA features resulting in lower benchmark accuracy only in the mixed feature case for ArabicMMLU and in the single feature case for AraDiCE-ArabicMMLU-EGY. Interestingly, in both cases the effect becomes no longer significant once more SEA features are introduced (three or more features for ArabicMMLU, two or more features for AraDiCE-ArabicMMLU-EGY). These results show that the effect of upstream bias on downstream performance varies substantially by model, although no model is completely unanimous across conditions in preferring CEA over SEA or exhibiting no preference. 

\section{Discussion}
The experiments in this paper have shown that the lack of SEA representation in text corpora used to train LLMs leads these models to view sentences with SEA features as less probable than their prestigious CEA counterparts, and that this upstream bias can lead to further impact downstream. It is therefore not the case that we can purely rely on the model's ability to generalize from massive amounts of text in ensuring that the model is equitable towards all Egyptian Arabic speakers.

This demonstration of bias leads us to assert the importance of not viewing dialect too reductively. Although ongoing efforts in Arabic NLP have aimed to face the diglossic nature of the language head-on, a lack of attention to the underlying makeup of major regional dialects implicitly favors the prestigious, ``default'' dialect. As was mentioned, this leads to an unfortunate feedback loop: the less that existing language technologies support SEA or other non-prestigious sub-dialects, the more likely SEA speakers are to avoid using their native dialect in digital spaces, resulting in less data on which to train and improve the model for that population. 

\section{Conclusion and Future Work}
Given the importance of accounting for sub-dialectal variation, the next step is identifying the existing varieties within each Arabic speaking country, the degree of variation, and whether this variation exists in the upstream representations and affects the downstream performance of LLMs for this dialect. This may require further efforts beyond available digital data, as some of these non-prestigious dialectal speakers could prefer not to position themselves as speakers of these marginalized dialects online, and accordingly we lack the necessary representative data. The present study has focused on dialectal features in an aggregate sense, without particular consideration for exactly which features drive upstream and downstream bias. Future work should aim to establish a causal relationship between individual features and model bias in order to develop a more comprehensive characterization of the problem.

\section*{Limitations}
The selected grammatical features distinguishing SEA and CEA were drawn from dialect surveys conducted in the 1980s~(\citealp{behnstedt:1985:DAA}; \citealp{khalafallah:2017:ADG}). As these descriptions were collected several decades ago, it is possible that some reported SEA specific features have changed in frequency or fallen out of use over time. To mitigate this limitation, we cross-referenced all selected features against a recently collected spoken Sa'idi Arabic corpus~\cite{eida:2025:BCS} and adjusted the feature set to reflect their observed frequency and usage. We then used regular expressions to automatically generate the MSA, CEA, SEA, and ungrammatical variants by replacing the selected grammatical features, followed by manual validation of the resulting items. Although every generated item was reviewed, some unintended substitutions or annotation errors may have remained despite our best efforts.

Another limitation of our work is that the models used in both experiments are open source models varying between 350M-13B parameters which are low in size in comparison to larger proprietary models. We understand that the performance on these larger proprietary models perform much better in multiple choice items, however, we were restricted to open source low-moderate sized models due to budgetary and computational constraints. 

\bibliography{custom}

\appendix
\onecolumn
\section{Probability Distributions}
\label{ProbabilityDistributions}
\begin{figure*}[h!]
\centering
\includegraphics[width=1\linewidth]{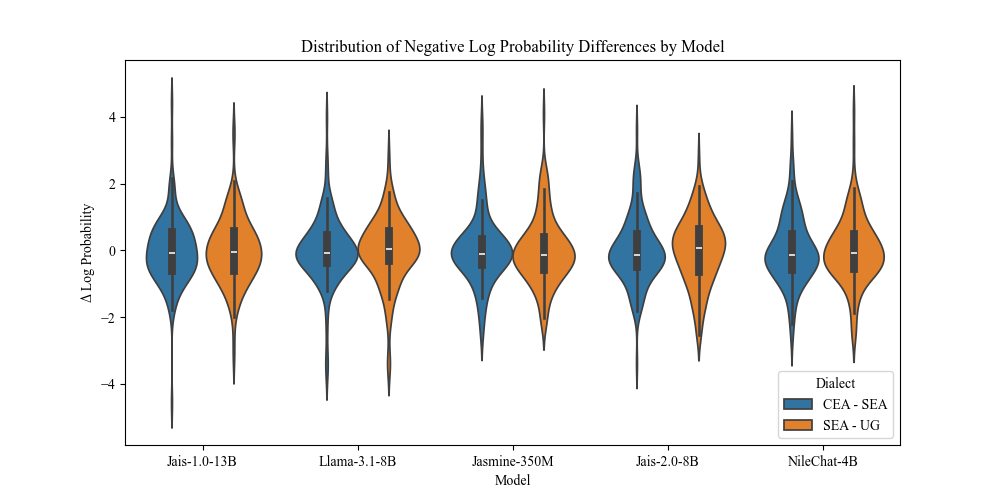}
\caption{Distribution of differences in negative log probabilities between Cairene Egyptian Arabic (CEA) and Sa'idi Egyptian Arabic (SEA), and between SEA and the ungrammatical baseline (UG) across five language models (Llama-3.1-8B, Jasmine-350M, and Jais-1-13B, Jais-2-8B, and NileChat-4B). Positive values indicate greater preference for the first variety in each comparison (CEA over SEA and SEA over UG), while negative values indicate greater preference for the second.}
\label{fig:deltaprobdistribution}
\end{figure*}

\begin{figure}[h!]
\centering
\includegraphics[width=1\linewidth]{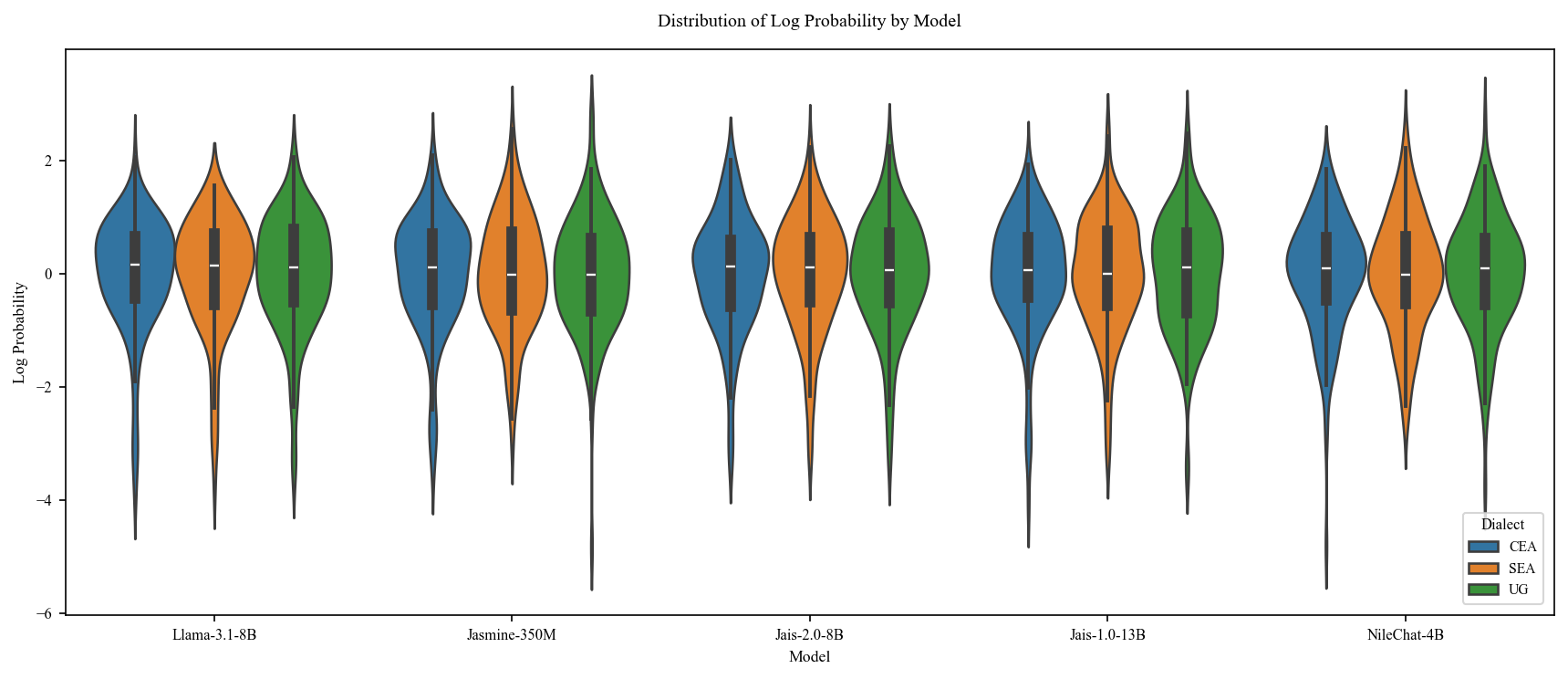}
\caption{Distribution of negative log probabilities for Cairene Egyptian Arabic (CEA), Sa'idi Egyptian Arabic (SEA), and ungrammatical inputs across five language models (Llama-3.1-8B, Jasmine-350M, and Jais-1-13B, Jais-2-8B, and NileChat-4B). Higher (positive) values indicate greater model preference.}
\label{fig:logprobabilitydistribution}
\end{figure}

\begin{figure}[h!]
\centering
\includegraphics[width=1\linewidth]{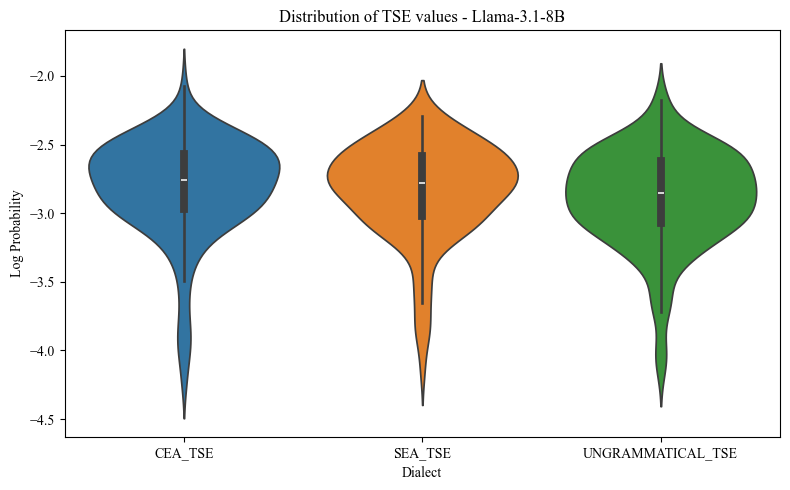}
\caption{Distribution of token-sequence log probabilities (TSE) for Cairene Egyptian Arabic (CEA), Sa'idi Egyptian Arabic (SEA), and ungrammatical for Llama-3.2-8B.}
\label{fig:LLAMAlogprobabilitydistribution}
\end{figure}

\vspace*{\fill}
\begin{figure}[h!]
\centering
\includegraphics[width=1\linewidth]{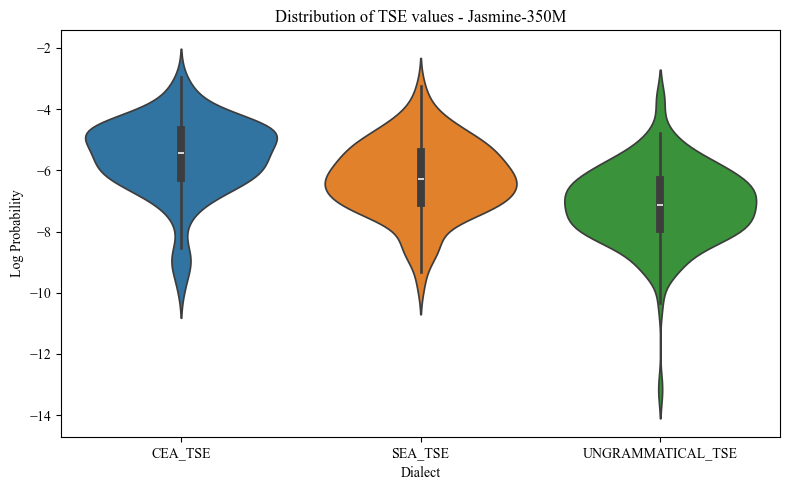}
\caption{Distribution of token-sequence log probabilities (TSE) for Cairene Egyptian Arabic (CEA), Sa'idi Egyptian Arabic (SEA), and ungrammatical for Jasmine-350M.}
\label{fig:JASMINElogprobabilitydistribution}
\end{figure}

\vspace*{\fill}

\begin{figure}[h!]
\centering
\includegraphics[width=1\linewidth]{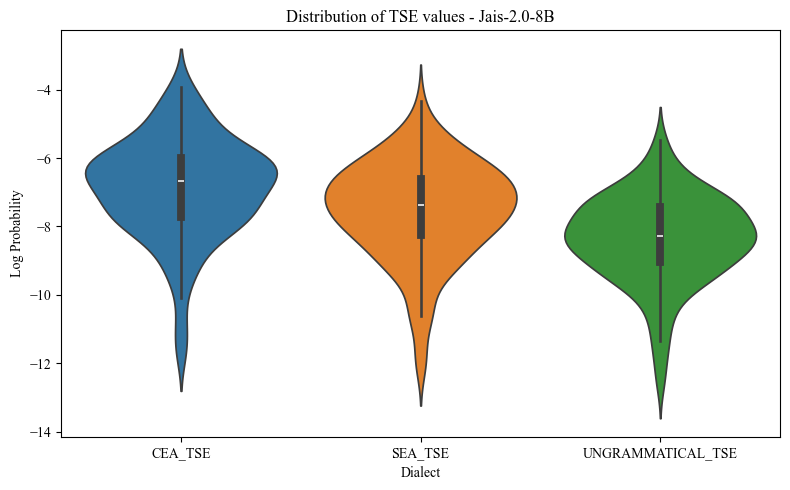}
\caption{Distribution of token-sequence log probabilities (TSE) for Cairene Egyptian Arabic (CEA), Sa'idi Egyptian Arabic (SEA), and ungrammatical for JAIS-2-8B}
\label{fig:JAIS2.0logprobabilitydistribution}
\end{figure}

\begin{figure}[h!]
\centering
\includegraphics[width=1\linewidth]{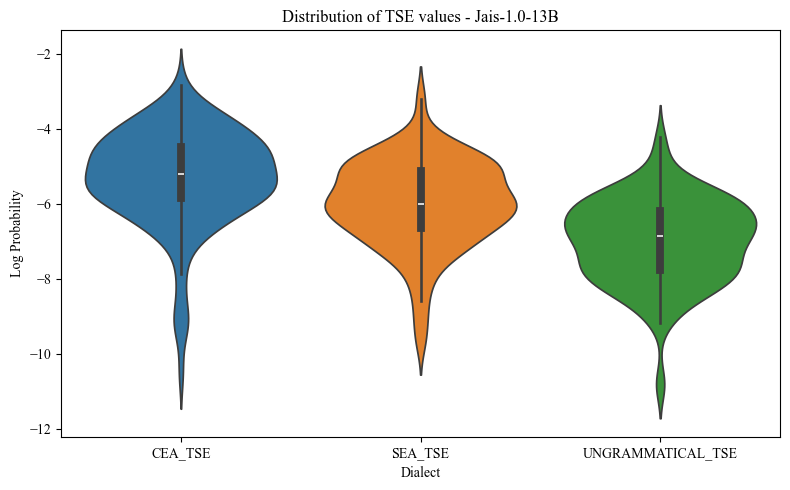}
\caption{Distribution of token-sequence log probabilities (TSE) for Cairene Egyptian Arabic (CEA), Sa'idi Egyptian Arabic (SEA), and ungrammatical for JAIS-1-13B.}
\label{fig:JAIS1.0logprobabilitydistribution}
\end{figure}

\begin{figure}[h!]
\centering
\includegraphics[width=1\linewidth]{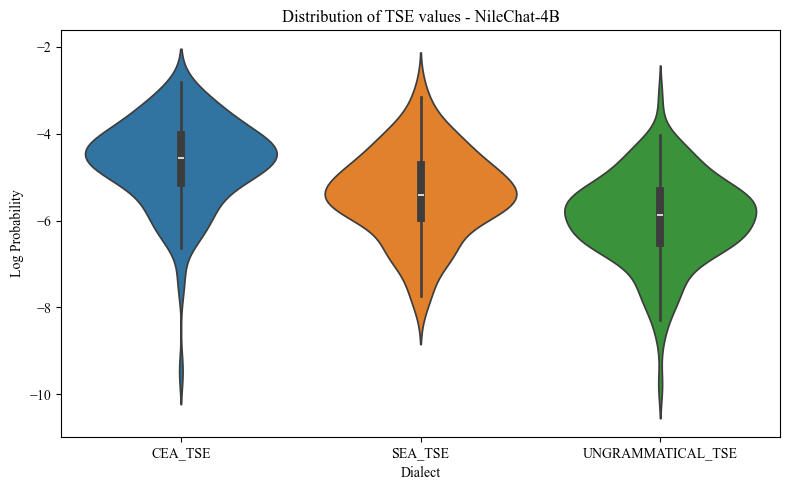}
\caption{Distribution of token-sequence log probabilities (TSE) for Cairene Egyptian Arabic (CEA), Sa'idi Egyptian Arabic (SEA), and ungrammatical in NileChat-4B.}
\label{fig:NileCHATlogprobabilitydistribution}
\end{figure}
\clearpage

\section{Grammatical Features}
\label{samples}
\begin{table*}[h!]
\centering
\includegraphics[width=1\linewidth]{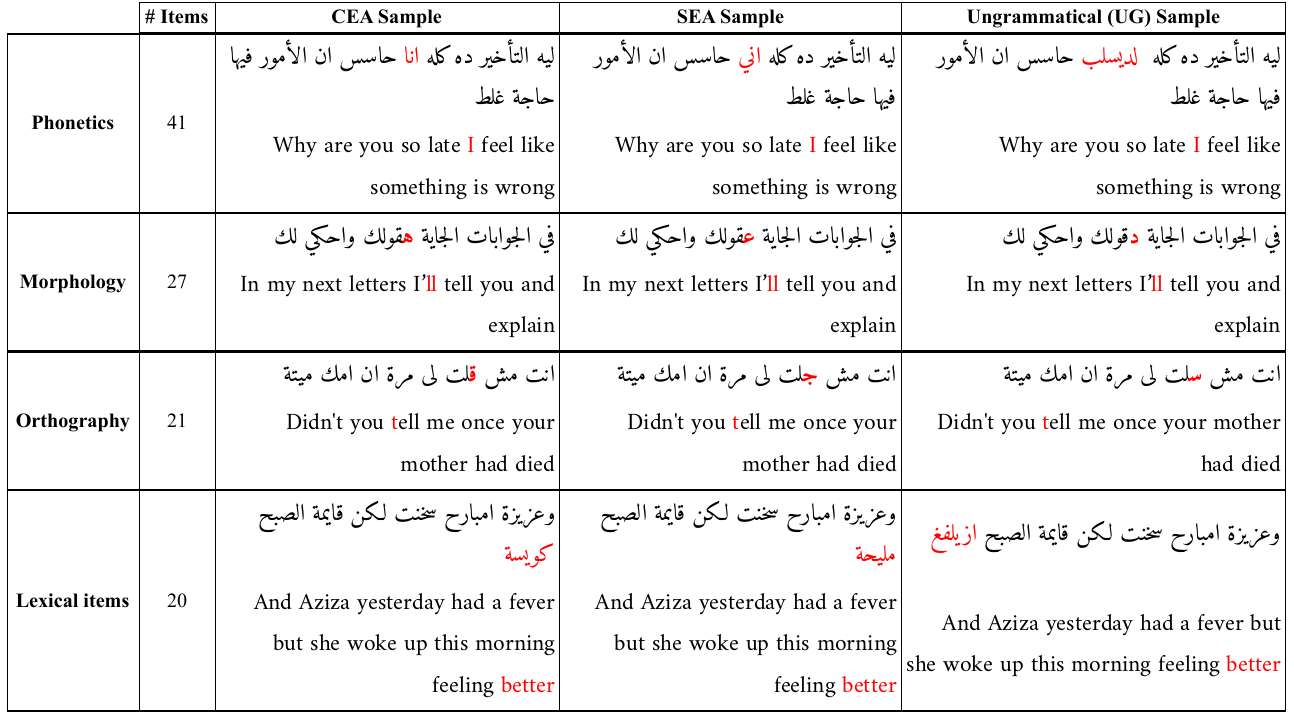}
\caption{Samples of CEA–SEA minimal pairs and ungrammatical controls (UG) used to evaluate dialectal preference in LLMs in Experiment 1. This table reports the number of items in each feature category and illustrates the manipulated marked grammatical dialectal features highlighted in red.}

\label{tab:minimalpairssamplesandstats}
\end{table*}

\begin{table*}[h!]
\centering
\includegraphics[width=1\linewidth]{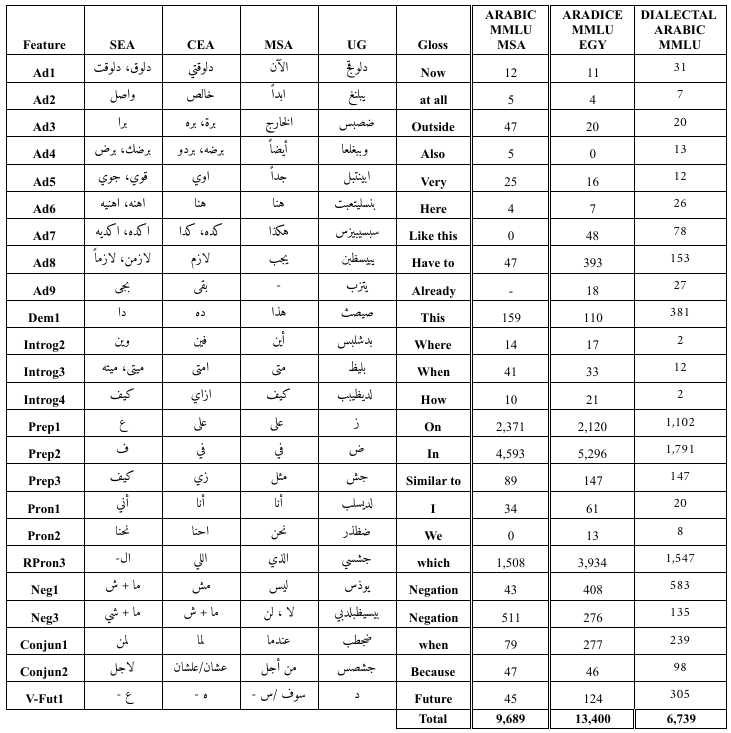}
\caption{SEA, CEA, MSA and UG variants of selected marked grammatical features across each dialect, followed by frequency of benchmark items containing each feature in the ArabicMMLU, AraDiCE-ArabicMMLU-EGY, and DialectalArabicMMLU benchmarks.}
\label{tab:listoffeatueresacrossbenchmarks}
\end{table*}

\end{document}